\documentclass[letterpaper, 10 pt, conference]{ieeeconf}
\IEEEoverridecommandlockouts
\usepackage{amsmath}
\usepackage{amssymb}
\usepackage{graphicx}
\usepackage{booktabs}
\usepackage{algorithm}
\usepackage{algorithmic}
\usepackage{cite}
\usepackage{url}
\usepackage[hidelinks]{hyperref}
\hypersetup{pdftitle={Goodhart in the Planner: Sampling-Based Control Exploits Rewards Learned from Human Preferences},
  pdfauthor={Soovadeep Bakshi, Apurba Bose}, pdfsubject={}, pdfkeywords={}}
\graphicspath{{figures/}}

\newtheorem{theorem}{Theorem}
\newtheorem{lemma}{Lemma}
\newtheorem{proposition}{Proposition}
\newtheorem{corollary}{Corollary}

\newcommand{\E}{\mathbb{E}}
\newcommand{\KL}{\mathrm{KL}}
\newcommand{\Jh}{\hat{J}}
\newcommand{\Cov}{\mathrm{Cov}}
\newcommand{\Var}{\mathrm{Var}}
\newcommand{\pf}{\emph{Proof:}~}
\newcommand{\qedhere}{\hfill$\square$}

\title{\LARGE \bf Goodhart in the Planner: Sampling-Based Control Exploits\\Rewards Learned from Human Preferences}
\author{Soovadeep Bakshi$^{1,\ast}$ and Apurba Bose$^{2,\ast}$%
\thanks{This work has been submitted to the IEEE for possible publication. Copyright may be transferred without notice,
after which this version may no longer be accessible.}%
\thanks{$^{\ast}$Equal contribution. $^{1}$Walker Department of Mechanical Engineering, The University of Texas at
Austin, Austin, TX, USA; corresponding author, {\tt\small soovadeep.bakshi@utexas.edu}. $^{2}$Independent researcher.}%
}

\begin{document}
\maketitle
\thispagestyle{empty}
\pagestyle{empty}

\begin{abstract}
Robots increasingly plan by searching against a reward model learned from human preferences. We measure what this
search does to the true objective. On HalfCheetah and Hopper with crowd-sourced labels, and on Meta-World drawer-open
and plate-slide with synthetic and blind human labels, harder MPPI search cuts true return by up to 76\% while
held-out label accuracy stays at 0.75--0.97; with human labels the drawer opens in 0 of 20 runs, against 10 of 10 for
the cloned policy the planner starts from. The effect persists with a learned dynamics model and an image scorer. We
prove that $\chi^2$-regularized MPPI at a temperature set by the scorer's error spread on the proposals is worst-case
optimal, and that no gain can be certified once that spread exceeds the scorer's own. Calibrated from 100 ranked
proposal pairs, this rule is the safest fix we test, keeping 86--94\% of the cloned policy's return on drawer-open
and plate-slide.
\end{abstract}

\section{Introduction}
\label{sec:intro}

A robot that plans against a learned reward is an optimizer pointed at a proxy. Reward models learned from human
preferences are now used inside model-predictive control \cite{krupa2025learning,elhasnaouy2026active,nabail2026ubp2},
and robot foundation models are scaled at test time by executing the one of many sampled actions that a learned
verifier likes best \cite{kwok2025robomonkey}. In language-model alignment, optimizing too hard against a learned reward
lowers the quality it was meant to measure \cite{gao2023scaling,coste2024ensembles,eisenstein2024helping}, and theory
explains when regularization toward a reference policy does and does not prevent it
\cite{huang2025chipo,huang2025bestofn,kwa2024catastrophic,laidlaw2025correlated}. Robot planners differ: the
optimizer is a sampling-based search whose strength is set by a temperature and a sample count, the regularizer is
implicit in the proposals, the plan is re-optimized at every step, and the labels number in the hundreds or
thousands. How much such a planner exploits its scorer has not been measured.

We study Model Predictive Path Integral control (MPPI) \cite{williams2017mppi,williams2017information} with proposals
around a policy cloned from the labelled data, the planning analogue of RLHF's reference model, and a Bradley--Terry
scorer trained from preference labels. With the true reward, search beats running the cloned policy on all five tasks of
Fig.~\ref{fig:lambda} (980 against 412 on HalfCheetah, 1,090 against 898 on the drawer; on hopper only with gentle
search). We ask how much of that gain a learned scorer keeps, planning on the true simulator so that every failure is
the scorer's, and check a learned model and an image scorer. Our contributions are:
\begin{itemize}
\item \textbf{Goodhart in the planner.} With real crowd-sourced labels \cite{yuan2024unirlhf} on the walkers, and
  synthetic and blind human labels on Meta-World \cite{yu2020metaworld}, harder search lowers true return below that of
  the cloned policy it starts from, while a perfect scorer gains or, on hopper, loses less. High agreement on decided
  human comparisons does not make planning safe.
\item \textbf{When search can be trusted.} Against an error spread $\sigma$ on the proposals, $\chi^2$-regularized MPPI
  at $\lambda\sqrt{\chi^2}=\sigma$ is worst-case optimal, no gain can be certified once $\sigma$ exceeds the scorer's own
  spread, and hard exponential search guarantees ever less as samples grow. The mathematics is elementary; what is new
  is bounding the error on the planner's own proposals, which held-out accuracy does not predict, and following it
  through the closed loop. Within each task, true return falls near one curve in $\sqrt{\KL}$ between the planner's
  weights and its proposals.
\item \textbf{Fixes, measured honestly.} Calibrated from 100 ranked proposal pairs, the worst-case-optimal rule is the
  safest fix we test, though cautious where search helps, and it cannot detect a bias the scorer shares with the
  labeller who calibrates it. Ensembles, $\chi^2$ weights, automatic temperatures and a guard each win somewhere and
  lose elsewhere; labelling the planner's own trajectories helps most.
\item \textbf{Best-of-$N$ is not safe either.} With a verifier from 500 synthetic pairs, selection falls below one
  noisy action from $N=16$.
\end{itemize}

\section{Related work}
\label{sec:related}

\emph{Reward overoptimization.} Gao et al.\ \cite{gao2023scaling} fitted how true quality falls with $\sqrt{\KL}$ from
the reference policy under best-of-$n$ and RL against a learned reward model. Ensembles
\cite{coste2024ensembles,eisenstein2024helping,zhai2026uncertainty}, weight averaging \cite{rame2024warm} and constraints
\cite{moskovitz2024confronting} mitigate it. Huang et al.\ showed that $\chi^2$ regularization gives guarantees KL
regularization cannot \cite{huang2025chipo} and analyzed best-of-$N$ with inference-time pessimism
\cite{huang2025bestofn}; our Lemma~\ref{lem:com} (Section~\ref{sec:theory}) centers and tightens their change-of-measure bound, and
Theorem~\ref{thm:chi2} turns it into a temperature MPPI can check from its own weights and a no-search threshold.
Reward hacking is defined in \cite{skalse2022defining,pan2022effects,laidlaw2025correlated}. These works study language
models or RL policies, not a planner in closed loop.

\emph{Preferences for robot control.} Preference-based reward learning
\cite{christiano2017deep,sadigh2017active,lee2021pebble,hejna2023fewshot} can pick up spurious features
\cite{tien2023causal}. Learned preference rewards are now optimized directly by MPC
\cite{krupa2025learning,elhasnaouy2026active} and by CEM-based planning in UBP2 \cite{nabail2026ubp2}, which plans
optimistically to collect labels; we study such planners at deployment.

\emph{Planning against learned models.} Model-based RL guards against exploiting learned dynamics with ensembles,
penalties \cite{chua2018pets,yu2020mopo,kidambi2020morel} or policy constraints
\cite{hansen2024tdmpc2,lin2025tdmpc2policy}; imperfect world models are exploitable \cite{bhamidipaty2026imperfect};
see also conservative MPPI \cite{yoon2026residual}. MPPI is KL-regularized by construction \cite{williams2017information},
and Tsallis-divergence MPPI \cite{wang2021tsallis} gives the $\chi^2$ rule. CriticHack \cite{luo2026critichack} evaluates
visual reward models under policy optimization.

\section{Setting}
\label{sec:setting}

\emph{Planner.} At each control step MPPI rolls out $N$ action sequences of length $H$ from the current state and
executes the first action of their weighted average. We draw them around the cloned policy's own closed-loop plan $U$:
$A_i=\mathrm{clip}(U+\epsilon_i)$, $\epsilon_i\sim\mathcal N(0,\sigma^2I)$, which plays the role of the reference
policy in RLHF. Each plan gets a learned return $\Jh(A_i)=\sum_{h<H}\hat r(o_h,a_h)$ and, for evaluation only, a true
return $J(A_i)$. Standard (exponential) MPPI uses $w_i\propto\exp(\Jh_i/\lambda)$, the solution of
$\max_w\sum_iw_i\Jh_i-\lambda\KL(w\Vert\mathrm{unif})$, RLHF's objective with $\beta=\lambda$. The $\chi^2$ rule of
Tsallis-divergence MPPI \cite{wang2021tsallis} uses $w_i\propto\max(0,\Jh_i-c)$, the solution with $\lambda\chi^2/2$
in place of $\lambda\KL$. Lower $\lambda$ means harder search; $\lambda\to0$ is best-of-$N$ over plans. Uniform weights
($\lambda=10^6$) give \emph{uniform averaging}, which executes the average of the clipped noisy proposals, not the
cloned policy's own action, so we also run \emph{the cloned policy directly}. Unless stated, $N=256$, $H=20$,
$\sigma=0.5$, and planning uses the true MuJoCo dynamics \cite{todorov2012mujoco}.

\emph{Scorers.} A two-layer MLP $\hat r(o,a)$ is trained on segment pairs with the Bradley--Terry likelihood
\cite{bradley1952rank,christiano2017deep}, with 10\% of pairs held out, and standardized per step; a tie (``can't
tell'') is a target of $\frac12$. Five seeds per label set give single scorers and a five-member ensemble
\cite{lakshminarayanan2017ensembles}.

\emph{Tasks and labels.} HalfCheetah-v5 and Hopper-v5 \cite{towers2024gymnasium} use D4RL medium-replay data
\cite{fu2020d4rl} and the Uni-RLHF crowd labels \cite{yuan2024unirlhf}: 2,000 pairs of 200-step segments that agree
with the true ranking on 73\% (cheetah) and 64\% (hopper) of pairs, with 10--15\% ties. Meta-World v3 drawer-open,
plate-slide and window-open \cite{yu2020metaworld} use mocap position control; the cloned policy is trained on 10
scripted demonstrations, the labelled data add 100 of its rollouts with action noise from 0 to 1, and 2,000 pairs of
50-step segments are labelled by a noisy Bradley--Terry teacher that agrees with the true ranking on 86--88\% of pairs;
smaller budgets use the first $K$ pairs. The authors also labelled pairs blind (two clips side by side in random
order, with ``can't tell''): drawer pairs 0--299 and 200--499 (97.6\% and 97.7\% agreement with the true ranking on
decided pairs, 18\% and 13\% ties; the two agreed on all 73 shared decided pairs) and plate-slide pairs 0--299
(97.2\%, 17\%). Episodes last 200 steps on the walkers and 150 on the arm, where we also count successes. Each cell
averages 10 runs (5 scorer seeds $\times$ 2 start states), and 20 for standard MPPI in the five main settings and for
the perfect scorer on the walkers. Code and results will be released.

\emph{Terms.} The \emph{scorer} is the learned reward model (the \emph{verifier} in best-of-$N$); the \emph{perfect
scorer} plans with the true reward. \emph{Gentle} and \emph{hard} search mean $\lambda=3$ and $\lambda=0.01$. An
\emph{oracle} ranks pairs by true return, which a deployed robot cannot do; in deployment a person would take its place.

\section{What the planner can and cannot certify}
\label{sec:theory}

Fix a state. Let $q$ give each of the $N$ proposals weight $1/N$ (the reference policy), $J$ and $\Jh$ be the true and
learned plan returns, and $e=\Jh-J$ the scorer's error; labels fix $\Jh$ only up to scale and offset, so $J$ is measured
in the scorer's units. A plan distribution $p\ll q$, such as MPPI's weights, has density $w=dp/dq$, the factor $w_i=Np_i$
by which plan $i$ is up-weighted. Let $\Delta(p)=\E_p[J]-\E_q[J]$ be the true gain of selecting with $p$, $\hat\Delta(p)$
the same under $\Jh$, and $s(p)=\sqrt{\chi^2(p\Vert q)}=(\tfrac1N\sum_i(w_i-1)^2)^{1/2}$ the distance moved: $s=0$ is
no search, and spreading the weight over the best $k$ plans gives $s=\sqrt{N/k-1}$ ($\approx16$ for the single best of
256). What matters is $\sigma_e=\mathrm{sd}_q(e)$, the spread of the scorer's mistake across the proposals: an error
shared by all of them changes no weights.

\begin{lemma}[Tight change of measure]\label{lem:com}
For every $p\ll q$, $|\Delta(p)-\hat\Delta(p)|\le\sigma_e\,s(p)$, with equality for an error of spread $\sigma_e$ that
overrates every plan in proportion to its extra weight $w-1$.
\end{lemma}
\emph{Proof:} $\hat\Delta-\Delta=\E_q[(w-1)(e-\E_qe)]$; apply Cauchy--Schwarz under $q$, with equality at
$e\propto w-1$.\qedhere

The bound sharpens the change-of-measure bound used for language models \cite{huang2025chipo,huang2025bestofn}: it
vanishes when the planner does not move, and no smaller bound holds. Against every error of spread at most $\sigma$,
the planner can therefore guarantee exactly $W_\sigma(p)=\hat\Delta(p)-\sigma s(p)$.

\begin{theorem}[Worst-case-optimal search]\label{thm:chi2}
(i) If $\sigma\ge\sigma_q(\Jh)$, no plan distribution guarantees a gain: $\max_pW_\sigma(p)=0$, attained by $p=q$.
(ii) If $\sigma<\sigma_q(\Jh)$, $W_\sigma$ is maximized by $\chi^2$ MPPI, $w=\max(0,(\Jh-c)/\lambda)$, at a
temperature $\lambda^\star$ with $\lambda^\star s(p_{\lambda^\star})=\sigma$, and the maximum is positive; along the
$\chi^2$ path, $W_\sigma$ rises until $\lambda^\star$ and falls after it. (iii) While no weight is clipped to zero,
$\hat\Delta(p_\lambda)=\sigma_q(\Jh)\,s$ and $\Delta(p_\lambda)=\rho_q\,\sigma_q(J)\,s$, where
$\rho_q=\mathrm{Corr}_q(J,\Jh)$.
\end{theorem}
\pf (i) By Cauchy--Schwarz, $\hat\Delta(p)=\E_q[(w-1)(\Jh-\E_q\Jh)]\le s(p)\,\sigma_q(\Jh)$, so
$W_\sigma(p)\le(\sigma_q(\Jh)-\sigma)\,s(p)\le0=W_\sigma(q)$.
(ii) $W_\sigma(p)=\E_q[w\Jh]-\E_q[\Jh]-\sigma\Vert w-1\Vert_{L^2(q)}$ is concave in $w$ on the convex set
$\{w\ge0,\ \E_qw=1\}$, so a point that satisfies the Karush--Kuhn--Tucker conditions is a global maximizer. At $w\ne1$
the gradient of $\Vert w-1\Vert$ is $(w-1)/s$; with a multiplier $\nu$ for $\E_qw=1$ and multipliers $\kappa\ge0$ with
$\kappa w=0$ for $w\ge0$, stationarity reads $\Jh-\sigma(w-1)/s-\nu+\kappa=0$, whose solution is
$w=\max(0,(\Jh-c)/\lambda)$ with $\lambda=\sigma/s$ and $c$ set by normalization: the $\chi^2$ weights at a temperature
with $\lambda s(p_\lambda)=\sigma$. Conversely, the $\chi^2$ weights at any $\lambda$ satisfy these conditions with
$\sigma=\lambda s(p_\lambda)$. Next, $p_\lambda$ maximizes $\E_q[w\Jh]-\tfrac\lambda2\E_q[(w-1)^2]$, so it maximizes
$\hat\Delta$ among all $p$ with $s(p)\le s(p_\lambda)$; call this maximum $g(s)$. $g$ is concave: if $w_1,w_2$ attain
$g(s_1),g(s_2)$, the mixture $tw_1+(1-t)w_2$ is feasible at $ts_1+(1-t)s_2$ by the triangle inequality and has
$\hat\Delta=tg(s_1)+(1-t)g(s_2)$. Along the path, $W_\sigma=g(s)-\sigma s$ is therefore concave in $s$, so it rises to
its maximizer and falls after; the maximizer moves to larger $s$ as $\sigma$ falls, so $\lambda s(p_\lambda)$ decreases
as $s$ grows. For $\lambda\ge\E_q\Jh-\min_i\Jh_i$ no weight is clipped, $w=1+(\Jh-\E_q\Jh)/\lambda$,
$s=\sigma_q(\Jh)/\lambda$ and $\lambda s=\sigma_q(\Jh)$; as $\lambda\to0$, $\lambda s\to0$ because $s\le\sqrt{N-1}$. The
product is continuous in $\lambda$, so $\lambda^\star$ exists for $\sigma>0$; for $\sigma=0$, $W_0=\hat\Delta$ is
maximized by putting all weight on the top-scored plans, the limit of the path as $\lambda\to0$. In both cases
$W_\sigma(p_{\lambda^\star})\ge W_\sigma(p_\lambda)=(\sigma_q(\Jh)-\sigma)s>0$ for any unclipped $\lambda$.
(iii) Unclipped, $w-1=(\Jh-\E_q\Jh)/\lambda$, so $\hat\Delta=\Var_q(\Jh)/\lambda=\sigma_q(\Jh)s$ and
$\Delta=\Cov_q(J,\Jh)/\lambda=\rho_q\sigma_q(J)s$.\qedhere

The stopping rule $\lambda\sqrt{\chi^2}=\sigma$ can be checked from the planner's own weights, and the $\chi^2$ geometry
comes from the error model: knowing only the error's spread makes $\sigma_e\sqrt{\chi^2}$ the exact worst case, while
sub-Gaussian errors would give a bound in $\sqrt{\KL}$, an assumption heavy tails break \cite{kwa2024catastrophic}.
\emph{Scale.} Putting $J$ in the scorer's units by the least-squares fit of $\Jh$ on $J$ gives
$\sigma_e=\sigma_q(\Jh)\sqrt{1-\rho_q^2}$, so the threshold of (i) becomes $\rho_q\le0$: a positive gain can be certified
exactly when the scorer correlates positively with the true return over the proposals (the theorem then guarantees the
sign of the gain, not its size). In scale-free form, $\chi^2$ MPPI with $\lambda s=\sigma$ does not lose for any true
return whose correlation with $\Jh$ over the proposals is at least $\sqrt{1-\sigma^2/\sigma_q(\Jh)^2}$. 

\begin{proposition}[The executed plan]\label{prop:exec}
MPPI executes the weighted average $\bar A_p=\E_p[A]$, not a plan drawn from $p$. If $J$ is $L$-smooth in the plan on
the action box, then $J(\bar A_p)\ge\E_p[J]-\tfrac L2\,\E_p\Vert A-\bar A_p\Vert^2$, so against every error of spread at
most $\sigma$, $J(\bar A_p)-\E_q[J]\ge W_\sigma(p)-\tfrac L2\,\E_p\Vert A-\bar A_p\Vert^2$. The last term vanishes as
the weights concentrate on one plan, and it can be dropped when $J$ is concave in the plan.
\end{proposition}
\pf $\bar A_p$ lies in the action box, which is convex. By $L$-smoothness, for every plan $A$,
$J(A)\le J(\bar A_p)+\nabla J(\bar A_p)^\top(A-\bar A_p)+\tfrac L2\Vert A-\bar A_p\Vert^2$. Taking $\E_p$, the linear
term vanishes, which gives the first inequality; the second follows from Lemma~\ref{lem:com}. For concave $J$, Jensen's
inequality gives $J(\bar A_p)\ge\E_p[J]$ directly.\qedhere

$L$ is unknown for contact-rich tasks, so we measured the averaging term $J(\bar A_p)-\E_p[J]$ directly: at least
$-0.05$ standard deviations except on window-open ($-0.15$).

\begin{corollary}[Hard exponential search]\label{prop:exp}
On $N$ samples with a unique top-scored plan, exponential MPPI with $\lambda\to0$ selects it and guarantees
$\max\Jh-\E_q\Jh-\sigma\sqrt{N-1}$. If the top score's margin grows more slowly than $\sqrt N$, as $\sigma_q(\Jh)\sqrt{2\log
N}$ does for Gaussian scores, this tends to $-\infty$ as $N$ grows; the guarantee of Theorem~\ref{thm:chi2} is never
negative.
\end{corollary}
\pf As $\lambda\to0$ the exponential weights tend to a point mass on the top plan $i^\star$: $w=N\,\mathbf 1[i=i^\star]$,
$s^2=\E_q[w^2]-1=N-1$ and $\hat\Delta=\Jh_{i^\star}-\E_q\Jh$. By Lemma~\ref{lem:com} the worst error is
$e-\E_qe=\sigma(w-1)/\sqrt{N-1}$, which is $\sigma\sqrt{N-1}$ on $i^\star$. If $\max\Jh-\E_q\Jh=o(\sqrt N)$, the
guarantee $\hat\Delta-\sigma\sqrt{N-1}\to-\infty$, while by Theorem~\ref{thm:chi2} $\max_pW_\sigma\ge W_\sigma(q)=0$.\qedhere

How many labels make search safe? For a linear reward the answer follows from Theorem~\ref{thm:chi2} and a known
estimation bound.

\begin{corollary}[Labels needed before search is safe]\label{cor:labels}
Let labels follow a correctly specified Bradley--Terry model with $r=\theta^{\star\top}\phi$, $\Vert\phi\Vert\le L$
summed over each segment and $\Vert\theta^\star\Vert\le B$, and let $\Jh$ come from the maximum-likelihood estimate
constrained to $\Vert\theta\Vert\le B$ on $n$ labelled pairs whose feature differences have second-moment matrix
$\Sigma_D$. With $\Sigma_q$ the covariance of plan features under $q$ and the coverage
$C_q=\lambda_{\max}\big((\Sigma_D+\mu I)^{-1/2}\Sigma_q(\Sigma_D+\mu I)^{-1/2}\big)$, with probability $1-\delta$,
$\sigma_e^2\le C^2C_q\big(\tfrac{d+\log(1/\delta)}{\gamma^2n}+\mu B^2\big)$ for any $\mu>0$, with the constants $C$
and $\gamma$ of \cite{zhu2023principled}. A positive gain can be certified once this is below $\sigma_q(\Jh)^2$, which
holds for $n>C^2C_q(d+\log(1/\delta))/\big(\gamma^2(\sigma_q(\Jh)^2-C^2C_q\mu B^2)\big)$ at any $\mu$ with
$C^2C_q\mu B^2<\sigma_q(\Jh)^2$.
\end{corollary}
\pf The scorer's error on a plan is $e=(\hat\theta-\theta^\star)^\top\Phi$, with $\Phi$ the plan's summed features, so
$\sigma_e^2=v^\top\Sigma_qv$ with $v=\hat\theta-\theta^\star$, which is at most $C_q\Vert v\Vert^2_{\Sigma_D+\mu I}$.
Lemma~3.1 of \cite{zhu2023principled} bounds $\Vert v\Vert_{\Sigma_D+\mu I}$ by
$C\sqrt{(d+\log(1/\delta))/(\gamma^2n)+\mu B^2}$ with probability $1-\delta$, with $\gamma=1/(2+e^{-LB}+e^{LB})$. A
positive gain can be certified when $\sigma_e<\sigma_q(\Jh)$, by Theorem~\ref{thm:chi2}; solving the bound for $n$ gives
the condition.\qedhere

The labels needed grow linearly with $C_q$, how much further the proposals vary than the labelled pairs did; wider
proposal noise and states the labelled data never reached both raise it.

\begin{proposition}[Where search starts to hurt]\label{prop:onset}
Let $p_\eta\propto q\,e^{\eta\Jh}$ with $\eta=1/\lambda$, and $d(\eta)=\sqrt{\KL(p_\eta\Vert q)}$.
(i) $\tfrac{d}{d\eta}\E_{p_\eta}[J]=\Cov_{p_\eta}(J,\Jh)$ and $\tfrac{d}{d\eta}\KL(p_\eta\Vert q)=\eta\Var_{p_\eta}(\Jh)$:
the true return of a plan drawn from the weights rises exactly while learned and true returns are positively correlated
under those weights. (ii) As $\eta\to0$, $\E_{p_\eta}[J]-\E_q[J]=\sqrt2\,\rho_q\,\sigma_q(J)\,d+O(d^2)$. (iii) The
executed plan satisfies $J(\bar A)-J(U)=\Delta(p)-c+\alpha$, with the price of noise $c=J(U)-\E_q[J]$ and the averaging
term $\alpha=J(\bar A)-\E_p[J]$, which vanishes as $\lambda\to0$ on $N$ samples.
\end{proposition}
\pf (i) $\log Z(\eta)=\log\E_q[e^{\eta\Jh}]$ has $\tfrac{d}{d\eta}\log Z=\E_{p_\eta}[\Jh]$, and
$\tfrac{d}{d\eta}\E_{p_\eta}[J]=\tfrac{d}{d\eta}\E_q[Je^{\eta\Jh}]/Z=\E_{p_\eta}[J\Jh]-\E_{p_\eta}[J]\E_{p_\eta}[\Jh]$.
With $\KL(p_\eta\Vert q)=\eta\E_{p_\eta}[\Jh]-\log Z(\eta)$,
$\tfrac{d}{d\eta}\KL=\E_{p_\eta}[\Jh]+\eta\Var_{p_\eta}(\Jh)-\E_{p_\eta}[\Jh]=\eta\Var_{p_\eta}(\Jh)$. Since $\KL$ is
increasing in $\eta$, the derivative along the distance has the same sign. (ii) At $\eta=0$,
$\E_{p_\eta}[J]=\E_q[J]+\eta\Cov_q(J,\Jh)+O(\eta^2)$ and $\KL=\tfrac{\eta^2}{2}\Var_q(\Jh)+O(\eta^3)$, so
$\eta=\sqrt2\,d/\sigma_q(\Jh)+O(d^2)$ and $\E_{p_\eta}[J]-\E_q[J]=\sqrt2\,\rho_q\sigma_q(J)\,d+O(d^2)$. (iii) The
decomposition is an identity; with a unique maximizer of $\Jh$, $p_\lambda$ tends to a point mass, so $\bar A$ tends to
that plan and $\alpha\to0$.\qedhere

Part (ii) makes $\sqrt{\KL}$ the natural axis, as Gao et al.\ \cite{gao2023scaling} found for language models, with a
slope set by how well the scorer ranks the proposals, not how well it fits its labels. Part (i) says that
exploitation begins where the scorer stops ranking the planner's own favourites correctly. Part (iii) separates
exploitation from a cost of sampling that a perfect scorer also pays; drawing one plan from the weights, as
inference-time pessimism does \cite{huang2025bestofn}, gives up the averaging term, which is why it does worse than
averaging on cheetah (167--265) and the drawer.

\begin{proposition}[Closed loop]\label{prop:loop}
With deterministic dynamics, let $V_t$ be the cloned policy's true return from step $t$ and $Q_t(s,a)$ that of taking
$a$ and then following it. If at each visited state $s_t$ the planner executes the first action $a_t$ of $\bar A_t$,
its episode return satisfies $R-V_0(s_0)=\sum_{t<T}[\Delta_t-c_t+\alpha_t+\eta_t]$, where $\Delta_t,c_t,\alpha_t$ are
the terms of Proposition~\ref{prop:onset}(iii) at $s_t$ and $\eta_t=Q_t(s_t,a_t)-V_t(s_t)-(J(\bar A_t)-J(U_t))$ is the
horizon gap. If the weights at every visited state are worst-case optimal for an error spread of at least that state's
$\sigma_e$, every $\Delta_t\ge0$.
\end{proposition}
\pf With deterministic dynamics $Q_t(s_t,a_t)=r_t+V_{t+1}(s_{t+1})$, with $V_T\equiv0$ (and $V\equiv0$ after
termination). Summing $Q_t(s_t,a_t)-V_t(s_t)$ over the episode telescopes to $\sum_tr_t-V_0(s_0)=R-V_0(s_0)$. Adding and
subtracting $J(\bar A_t)-J(U_t)$ in each term and splitting it as in Proposition~\ref{prop:onset}(iii) gives the
identity, along every realization of the proposal noise. If the weights $p_t$ maximize $W_\sigma$ for some
$\sigma\ge\sigma_e(s_t)$, then by Lemma~\ref{lem:com} $\Delta_t\ge\hat\Delta_t-\sigma_e(s_t)\,s(p_t)\ge W_\sigma(p_t)\ge
W_\sigma(q)=0$.\qedhere

The other terms still depend on the scorer, which sets the weights and the states visited; Theorem~\ref{thm:chi2}
controls the selection terms, given the error spread on the visited states, and we measure the rest.

\section{Experiments}
\label{sec:exp}

\subsection{Harder search lowers true return}
\label{sec:exp-lambda}

\begin{figure*}[t]
\centering
\includegraphics[width=\textwidth]{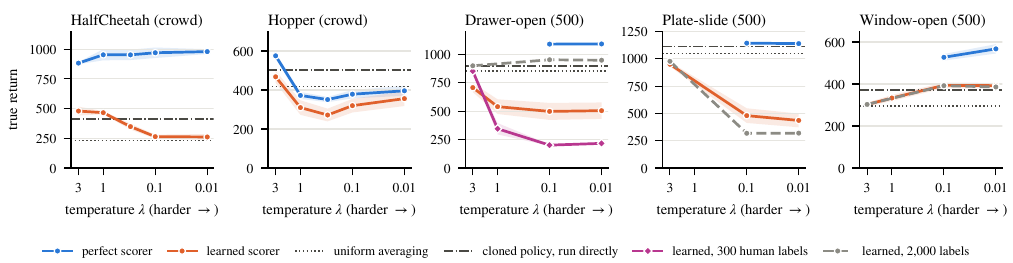}
\vspace{-6mm}
\caption{True return against MPPI temperature (harder search to the right), mean $\pm1$ standard error over 10--20
runs. A learned scorer loses return under harder search on every task except window-open; the perfect scorer keeps or
improves its return except on hopper; 300 human labels are exploited fastest.}
\label{fig:lambda}
\end{figure*}

Fig.~\ref{fig:lambda} shows the main effect. With the crowd scorer, true return falls under harder search on both
walkers (cheetah 480 to 262, hopper 468 to 356), while the perfect scorer improves on cheetah (884 to 980). On hopper
even the perfect scorer loses (575 to 396): at $\lambda\le1$ the hopper falls before the episode ends in most runs, a
horizon gap that the 20-step return cannot see. On the arm, where the cloned policy is already near the optimum,
harder search hurts with 500 synthetic labels: true return falls from 707 to 504 on the drawer, which stays shut in 9
of 20 runs, and from 950 to 436 on plate-slide, while uniform averaging and the cloned policy succeed every time. With
2,000 labels search becomes safe on the drawer (899 to 948) but not on plate-slide (977 to 319).

Human labels are the starkest case. With one author's 300 blind drawer labels the gentlest search matches uniform
averaging (851, 20 of 20 opened), but at every temperature at or below 0.1 the drawer stays shut in all 20 runs, and
both authors' 500 labels fare no better. On plate-slide human labels are exploited less than synthetic ones on the same
pairs (1,066 against 499 at hard search), though both authors' 500 still fall to 681. Search is not always exploited:
on window-open it helps at every label budget (295 without search, 375--392 at hard search), and button-press is a null
case. Hard search still loses with other horizons and proposal noise, except on cheetah with smaller noise.

\begin{figure}[t]
\centering
\includegraphics[width=\columnwidth]{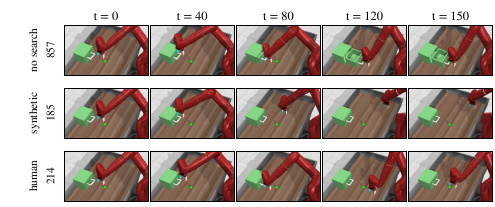}
\caption{One drawer start under three planners (true returns). Uniform averaging opens the drawer. Standard MPPI
($\lambda=0.01$) on the 500-label scorer lifts the hand into a part of the workspace the data never visited; on the
human-label scorer ($\lambda=0.1$) it pulls in front of the handle without touching it.}
\label{fig:film}
\end{figure}

\emph{Why accurate human labels are exploited.} The human labels agree with the true ranking on 97.6\% of decided pairs
and give the best held-out accuracy of any scorer we trained (0.97), yet they make the worst planning target. Controls
on the same 300 pairs point to the 55 pairs the person could not tell apart: ordering every pair by true return, or
dropping 55 random pairs, lets hard search open the drawer in 6 of 10 runs, while with only the decided pairs in true
order it never opens. What the planner needs is the order of those close pairs. In recorded episodes
(Fig.~\ref{fig:film}) the arm parks in an unvisited corner that the scorer rates highly (synthetic labels) or pulls about
10\,cm short of the handle (human labels).

\subsection{True return follows the distance moved}
\label{sec:exp-law}

\begin{figure*}[t]
\centering
\includegraphics[width=\textwidth]{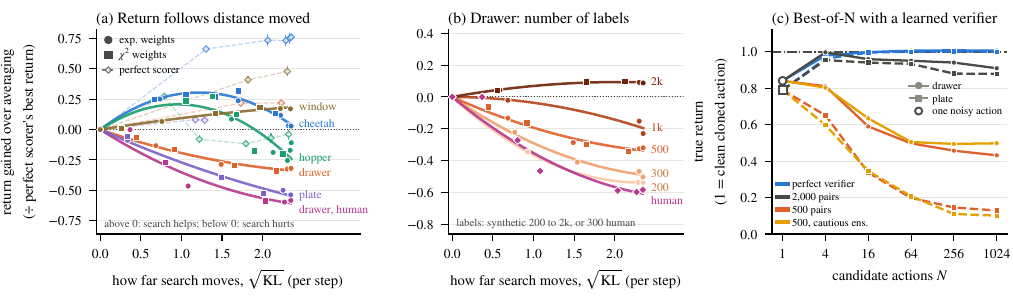}
\vspace{-6mm}
\caption{(a) True return gained over uniform averaging, as a fraction of the perfect scorer's best return, against
the distance $d=\sqrt{\KL}$ (arm tasks: 500 labels). Circles: exponential weights; squares: $\chi^2$ weights; one point
per temperature; curves: fits of $a\,d-b\,d^2$; open diamonds: perfect scorer ($\pm1$ standard error). (b) The drawer
for five label budgets and the human labels. (c) Best-of-$N$ on the drawer (solid) and plate-slide (dashed), relative
to the clean cloned action (dash-dot); $N=1$ (open markers) is one noisy action.}
\label{fig:kl}
\end{figure*}

Fig.~\ref{fig:kl}a plots every single-scorer MPPI cell against $d$. Within each task the cells fall near one curve
$\Delta J=a\,d-b\,d^2$ through uniform averaging ($R^2=0.79$--$0.99$), and the sign of the initial slope $a$ is clear on
every task. A curve fitted on the exponential cells alone predicts the $\chi^2$ cells to within 16--57\% of their
spread: at equal distance the two rules give similar true return, so $\chi^2$ weights help mainly by moving less. The
walkers first gain and then lose (peaks near $d=1.2$ and 1.0); the drawer and plate-slide lose from the start, and
window-open gains throughout. The perfect scorer at the same distances separates exploitation from the planner's own
cost: on cheetah, the drawer and plate-slide the gap opens from the first step and reaches 590--720 points at hard
search. On the drawer (Fig.~\ref{fig:kl}b) the initial slope grows roughly linearly in $\log n$ with the label count
and crosses zero near 1,200 labels; the human labels sit below the synthetic labels on the same 300 pairs, though the
intervals overlap. On plate-slide, 2,000 labels do not change the sign.

\subsection{Two ways to be exploited, and why held-out checks do not warn}
\label{sec:exp-mech}

\begin{figure*}[t]
\centering
\includegraphics[width=\textwidth]{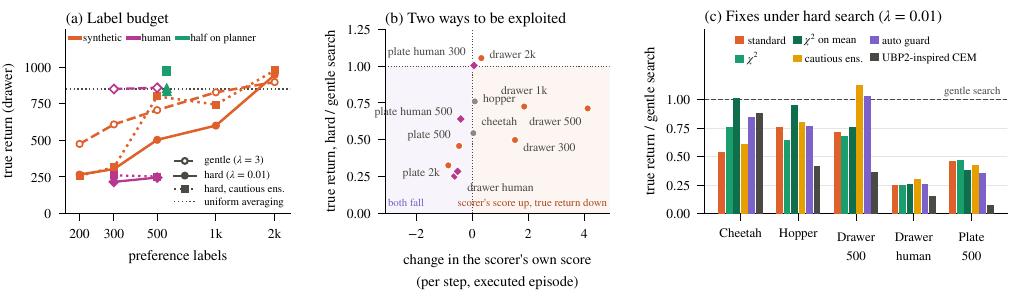}
\vspace{-6mm}
\caption{(a) Drawer: true return against label budget at gentle and hard search; green: 500 labels, half on the
planner's own trajectories (Section~\ref{sec:exp-fix}). (b) Change in the scorer's estimate of the executed episode from
$\lambda=3$ to $\lambda=0.01$, against the ratio of true returns; diamonds: human labels. (c) Each fix at
$\lambda=0.01$, relative to standard MPPI at $\lambda=3$.}
\label{fig:mech}
\end{figure*}

Fig.~\ref{fig:mech}b separates two patterns by the sign of the change in the scorer's own estimate of the episode the
robot executes. On the drawer with synthetic labels the estimate rises sharply as true return falls, which we read as
the planner finding states the scorer overrates, the extremal case of Manheim and Garrabrant
\cite{manheim2018categorizing}. On plate-slide and with human labels the estimate \emph{falls}, which we read as chasing
plans that look good at planning time and do not hold up once the planner re-plans, the optimizer's curse
\cite{smith2006optimizer} in closed loop. Leaving the labelled data alone does not make a scorer exploitable: on
window-open most states are off the data even without search, and search still helps.

Why do the usual checks not warn? We probed the planner's 256 proposals along the cloned policy's episodes, scored by
the learned scorer and the true reward. The initial slope that Proposition~\ref{prop:onset}(ii) predicts matches the
measured one to within 6\% in eleven of the thirteen settings, and the correlation $\rho_q$ that sets it is unrelated
to the held-out checks: the 500-label drawer scorer has held-out accuracy 0.88 but ranks the planner's proposals with
$\rho_q=0.07$, while the hopper scorer, with a per-step correlation of only 0.08 on the labelled data, ranks them with
$\rho_q=0.66$. Perturbing the cloned plan costs 0.7--2.2 standard deviations of true plan return, and the scorers
recover only part of that by selection; on window-open the cost is only 0.33 and selection recovers 0.7--1.0, which is
why search helps there. A scorer can therefore be checked only where it will be optimized: on the planner's own
proposals and on states the cloned policy never visited. On the states hard search visits, the scorer ranks the
proposals worse than on the cloned policy's states in every exploited setting but cheetah. Summed over a drawer
episode, the selection gains of Proposition~\ref{prop:loop} are $-151$ with 500 labels and $-373$ with human labels,
against $+632$ with the perfect scorer; on plate-slide they stay positive, and the loss is in the horizon gap.

\begin{algorithm}[t]
\caption{Calibrated $\chi^2$ MPPI}
\label{alg:calib}
\begin{algorithmic}
\STATE \textbf{Once:} rank $M=100$ pairs of the planner's proposals (by an oracle or a person); let $c$ be the
share the scorer orders the same way; $\hat\rho\leftarrow\sin(\pi(c-\frac12))$ (or its tie-corrected value);
$\kappa\leftarrow\sqrt{1-\max(0,\hat\rho)^2}$
\FOR{each control step}
\STATE draw $A_i=\mathrm{clip}(U+\epsilon_i)$, $i\le N$; score each by the five-scorer mean $\Jh_i$
\STATE $\sigma\leftarrow\kappa\cdot\mathrm{sd}_i(\Jh_i)$
\IF{$\hat\rho\le0$}
\STATE $w_i\leftarrow1$ for all $i$ (no search)
\ELSE
\STATE find $\lambda$ with $\lambda\,s(p_\lambda)=\sigma$ by bisection; $w_i\leftarrow\max(0,(\Jh_i-\tau)/\lambda)$,
$\tau$ such that $\bar w=1$
\ENDIF
\STATE execute the first action of $\frac1N\sum_iw_iA_i$
\ENDFOR
\end{algorithmic}
\end{algorithm}

The same proposals test Theorem~\ref{thm:chi2}. Given the true error spread, the worst-case-optimal rule does not lose
at any of the 2,042 states and gains 0.05--0.53 standard deviations on average; standard MPPI at hard search gains
more on average in ten of thirteen settings but loses 1.6--8.0 at its worst state, which a closed-loop planner has many
chances to meet. The unknown is the error spread, which the ensemble's disagreement underestimates by a factor of up to
2.8: scorers trained on the same labels share their mistakes. Algorithm~\ref{alg:calib} instead calibrates it from the
share of 100 pairs of the planner's own proposals that the scorer orders correctly, using the Gaussian relation between
concordance and correlation \cite{kruskal1958ordinal}. What the estimate must get right follows from
Lemma~\ref{lem:com} and Theorem~\ref{thm:chi2}.

\begin{corollary}[Calibrated rule]\label{cor:calib}
(i) At a state whose proposals have $\rho_s=\mathrm{Corr}_q(J,\Jh)>0$, the rule with
$\sigma=\sigma_q(\Jh)\sqrt{1-\hat\rho_+^2}$, $\hat\rho_+=\max(0,\hat\rho)$, does not lose ($\Delta\ge0$) whenever
$\hat\rho\le\rho_s$, and it does not search when $\hat\rho\le0$. (ii) Let a labeller rank pairs by $J+u$, with $J$ in
the scorer's units, $\Jh=J+e$, and $e$ and $u$ uncorrelated with $J$. If $(\Jh,J+u)$ is bivariate Gaussian under $q$,
the concordance $c$ of the labeller with $\Jh$ on two independent proposals satisfies
$\sin(\pi(c-\frac12))=\rho_u=\mathrm{Corr}_q(\Jh,J+u)$, and $\rho_u>\rho_q$ exactly when
$\Cov_q(e,u)>\sigma_q(J)\big(\sqrt{\sigma_q(J)^2+\sigma_u^2}-\sigma_q(J)\big)$, about $\sigma_u^2/2$ for a small
labelling error; with $\Cov_q(e,u)=0$, $\rho_u<\rho_q$ whenever $\sigma_u>0$.
\end{corollary}
\pf (i) By the scale argument $\sigma_e=\sigma_q(\Jh)\sqrt{1-\rho_s^2}\le\sigma$, so the rule's weights $p$ maximize
$W_\sigma$ and $\Delta(p)\ge\hat\Delta(p)-\sigma_es(p)\ge W_\sigma(p)\ge W_\sigma(q)=0$ by Lemma~\ref{lem:com}; if
$\hat\rho\le0$ then $\sigma=\sigma_q(\Jh)$ and Theorem~\ref{thm:chi2}(i) gives $p=q$. (ii) For a bivariate Gaussian, two
independent draws are concordant with probability $\frac12+\arcsin(\rho_u)/\pi$ \cite{kruskal1958ordinal}; and
$\rho_u=\big(\sigma_q(J)^2+\Cov_q(e,u)\big)\big/\big(\sigma_q(\Jh)\sqrt{\sigma_q(J)^2+\sigma_u^2}\big)$ and
$\rho_q=\sigma_q(J)/\sigma_q(\Jh)$.\qedhere

Independent labelling noise therefore makes the rule more cautious, while a mistake the labeller shares with the
scorer can make it trust the scorer too much. Ranked by an oracle, it loses at most 0.22 standard
deviations at its worst state in twelve of thirteen settings and keeps 47--93\% of the true-spread rule's gain where
that gain exceeds 0.2; a lower confidence bound on $c$ cuts the loss further, at a cost in gain. Mistakes the labeller
shares with the scorer, sampling noise and pooling over states can each put $\hat\rho$ above a state's own, so the rule
is an empirical safeguard, not a per-state certificate.

\subsection{What helps, and where}
\label{sec:exp-fix}

\begin{table*}[t]
\centering
\caption{True return at hard search, mean $\pm$ standard error over 10 runs per cell ($^\dagger$\,20); successes over
runs in parentheses on the arm. $^\ddagger$\,Mean $\pm$ standard deviation across five independent 100-pair
calibration sets ranked by an oracle, drawn on start states the evaluation never uses, each calibrating its own
controller. Bold: best learned-scorer method per column.}
\label{tab:fixes}
\footnotesize
\begin{tabular}{lccccc}
\toprule
& Cheetah & Hopper & Drawer 500 & Drawer human & Plate 500 \\
\midrule
Cloned policy, run directly & 412\,{\scriptsize$\pm$114} & 502\,{\scriptsize$\pm$32} & 898\,{\scriptsize$\pm$1}\,(10/10) & 898\,{\scriptsize$\pm$1}\,(10/10) & 1111\,{\scriptsize$\pm$5}\,(10/10) \\
Uniform averaging (no search) & 233\,{\scriptsize$\pm$78} & 418\,{\scriptsize$\pm$38} & 852\,{\scriptsize$\pm$1}\,(10/10) & 852\,{\scriptsize$\pm$1}\,(10/10) & 1051\,{\scriptsize$\pm$6}\,(10/10) \\
Perfect scorer, $\lambda=0.01$ & 980\,{\scriptsize$\pm$21}$^\dagger$ & 396\,{\scriptsize$\pm$26}$^\dagger$ & 1090\,{\scriptsize$\pm$7}\,(10/10) & 1090\,{\scriptsize$\pm$7}\,(10/10) & 1139\,{\scriptsize$\pm$7}\,(10/10) \\
Learned scorer, gentle ($\lambda=3$) & 480\,{\scriptsize$\pm$38}$^\dagger$ & 468\,{\scriptsize$\pm$31}$^\dagger$ & 707\,{\scriptsize$\pm$41}$^\dagger$\,(18/20) & 851\,{\scriptsize$\pm$4}$^\dagger$\,(20/20) & 950\,{\scriptsize$\pm$21}$^\dagger$\,(20/20) \\
\midrule
\multicolumn{6}{l}{\emph{Learned scorer, hard search ($\lambda = 0.01$)}} \\
Standard (exp.\ weights) & 262\,{\scriptsize$\pm$25}$^\dagger$ & 356\,{\scriptsize$\pm$39}$^\dagger$ & 504\,{\scriptsize$\pm$74}$^\dagger$\,(11/20) & 216\,{\scriptsize$\pm$10}$^\dagger$\,(0/20) & 436\,{\scriptsize$\pm$65}$^\dagger$\,(15/20) \\
$\chi^2$ weights & 364\,{\scriptsize$\pm$23} & 303\,{\scriptsize$\pm$49} & 480\,{\scriptsize$\pm$96}\,(5/10) & 211\,{\scriptsize$\pm$9}\,(0/10) & 448\,{\scriptsize$\pm$80}\,(7/10) \\
Cautious ensemble (mean $-$ sd) & 295\,{\scriptsize$\pm$30} & 374\,{\scriptsize$\pm$30} & 801\,{\scriptsize$\pm$51}\,(10/10) & 261\,{\scriptsize$\pm$42}\,(1/10) & 406\,{\scriptsize$\pm$68}\,(8/10) \\
$\chi^2$ on ensemble mean & 485\,{\scriptsize$\pm$25} & 448\,{\scriptsize$\pm$56} & 538\,{\scriptsize$\pm$85}\,(6/10) & 219\,{\scriptsize$\pm$13}\,(0/10) & 360\,{\scriptsize$\pm$75}\,(7/10) \\
Automatic guard & 409\,{\scriptsize$\pm$31} & 361\,{\scriptsize$\pm$60} & 729\,{\scriptsize$\pm$77}\,(9/10) & 222\,{\scriptsize$\pm$17}\,(0/10) & 335\,{\scriptsize$\pm$72}\,(6/10) \\
\midrule
\multicolumn{6}{l}{\emph{Learned scorer, adaptive temperature or a different optimizer}} \\
Automatic $\lambda$ ($\chi^2$, $c=0.5$) & \textbf{521}\,{\scriptsize$\pm$20} & \textbf{569}\,{\scriptsize$\pm$1} & 650\,{\scriptsize$\pm$60}\,(7/10) & 262\,{\scriptsize$\pm$23}\,(0/10) & 662\,{\scriptsize$\pm$61}\,(10/10) \\
Worst-case-optimal $\lambda$ (Thm.~\ref{thm:chi2}, $\kappa=1$) & 460\,{\scriptsize$\pm$52} & 569\,{\scriptsize$\pm$2} & 575\,{\scriptsize$\pm$76}\,(8/10) & 231\,{\scriptsize$\pm$13}\,(0/10) & 699\,{\scriptsize$\pm$58}\,(10/10) \\
Worst-case-optimal $\lambda$, 100 oracle pairs (Alg.~\ref{alg:calib}) & 357\,{\scriptsize$\pm$86}$^\ddagger$ & 539\,{\scriptsize$\pm$12}$^\ddagger$ & \textbf{840}\,{\scriptsize$\pm$16}$^\ddagger$\,(50/50) & \textbf{852}\,{\scriptsize$\pm$0}$^\ddagger$\,(50/50) & \textbf{959}\,{\scriptsize$\pm$86}$^\ddagger$\,(50/50) \\
CEM, optimistic (UBP2-inspired) & 423\,{\scriptsize$\pm$19} & 195\,{\scriptsize$\pm$43} & 260\,{\scriptsize$\pm$88}\,(2/10) & 129\,{\scriptsize$\pm$2}\,(0/10) & 68\,{\scriptsize$\pm$1}\,(0/10) \\
\bottomrule
\end{tabular}
\end{table*}

Table~\ref{tab:fixes} and Fig.~\ref{fig:mech}c compare the fixes at hard search; none wins everywhere. Among fixed
temperatures, $\chi^2$ weights on the mean of five scorers do best on the walkers, and the cautious ensemble (mean minus
one standard deviation \cite{coste2024ensembles}) helps less. On the drawer with synthetic labels the ranking reverses:
the cautious ensemble restores 801 (10/10) while $\chi^2$ weights do not help. This fits the two patterns above: where
the planner reaches states the scorers overrate, they disagree and the penalty finds them; where it chases noise, they
agree, and only gentler search helps. Automatic temperatures are the most even across tasks; with human labels every
hard-search fix fails.

Calibrated from 100 oracle pairs, Algorithm~\ref{alg:calib} is the safest fix. Each of five independent calibration
sets per task, drawn on start states the evaluation never uses, calibrates its own controller. On the drawer and
plate-slide every controller succeeds in all its runs and keeps on average 86--94\% of the cloned policy's return,
against 611 (8/10) and 351 (6/10) for hard search on the same five scorers. With human labels every set declines to
search. The rule also gains on hopper with every set, lands between uniform averaging and the cloned policy on cheetah,
and pays for caution on window-open, where search helps.

\emph{Calibrating from people.} Calibrated instead from one author's comparisons of 100 such pairs, the rule trusts the
scorer trained on that author's labels and opens the drawer in only 5 of 10 runs (574); the other author's comparisons
leave it barely searching (845, 10/10). Both people called about 70\% of pairs ties, which inflates the concordance on
the rest: if a person ties pairs whose perceived returns differ by less than a threshold \cite{rao1967ties}, the decided
pairs are the clearer ones. Correcting $\hat\rho$ for the observed share of ties under the same Gaussian model, the
first author's calibration scores 822 (10/10). Calibration against the scorer's own labeller cannot detect a bias they
share; an independent evaluator, the tie correction or the lower bound limits the damage.

A UBP2-inspired optimistic CEM planner (CEM as in \cite{nabail2026ubp2} on the five-scorer mean plus one standard
deviation) is the strongest optimizer with the perfect scorer and the most exploited on the arm (plate-slide 68--79
with no successes); its collapse persists without optimism and at matched budget.

\emph{Labelling where the planner goes.} The scorer is exploited where its labels did not reach, so, as active
preference learning for MPC does \cite{elhasnaouy2026active}, we spend half of a 500-label drawer budget on the planner
itself: 250 new pairs from MPPI episodes planned with a scorer trained on the first 250 offline pairs. Hard search then
scores 834 (8/10) instead of 504 (11/20), and the cautious ensemble 975 (10/10), level with 2,000 offline labels and
above the cloned policy (898; Fig.~\ref{fig:mech}a).

\subsection{Best-of-$N$, a learned dynamics model and an image scorer}
\label{sec:exp-bon}

We test best-of-$N$ action selection \cite{kwok2025robomonkey} with a state-based verifier trained on 500 or 2,000
synthetic pairs over perturbed actions. The $N$ candidates are noisy copies of the cloned action, so the reference is
one noisy action (755 on the drawer, 878 on plate-slide), and a perfect verifier recovers the clean action's return
(898 and 1,111) for every $N$ (Fig.~\ref{fig:kl}c). With 500 pairs selection falls well below one noisy action from
$N=16$ on both tasks (530 and 375), and on plate-slide already at $N=4$; with 2,000 pairs it beats one noisy action at
every $N$.

Neither the simulator nor the state scorer drives the effect. Planning through a five-member MLP ensemble fit to the
logged transitions \cite{chua2018pets}, the perfect scorer still gains (cheetah 688, drawer 1,013 at hard search,
against 412 and 898 for the cloned policy), while the learned scorer still loses under harder search (cheetah 377 to
255, drawer 842 to 650). A CNN scorer on $64\times64$ camera frames, trained on the same 500 drawer labels, ranks
held-out pairs as well as the state scorer (0.87 against 0.88), yet hard search against it opens the drawer in 1 of 10
runs (283, against 791 at $\lambda=3$; state scorer 504).

\section{Limitations and conclusion}
\label{sec:conclusion}

Our tasks are simulated, our scorers are small networks, the learned-model and image checks cover only cheetah and the
drawer, the human labels come from two people on two tasks, and most cells have 10 runs. Most calibration pairs are
ranked by true return, an oracle a deployed robot lacks, apart from two people's 100 comparisons; the best-of-$N$
verifiers use synthetic labels. Our guarantee covers only the selection terms of the closed loop, and the calibrated
rule is an empirical safeguard. The distance curves and the label trend summarize the range we tested, and the two
exploitation patterns are our reading, not proven mechanisms. Window-open, button-press and two Adroit hand tasks are
null or limit cases.

In practice, temperature, sample count and elite fraction are an optimization budget against the scorer, set by its
quality \emph{on the planner's own proposals}. A calibrated temperature was the safest default, given an evaluator
independent of the labels; labelling where the planner goes helped most.

\section*{Acknowledgment}
In line with IEEE policy, generative AI (Claude, Anthropic) was used, under the authors' direction, for the
literature search (Sections~\ref{sec:intro} and~\ref{sec:related}), the experiment code, figures and tables
(Sections~\ref{sec:setting} and~\ref{sec:exp}), the derivations in Section~\ref{sec:theory}, and the writing and
editing of the text. The authors reviewed all of it and take full responsibility for the content.

\enlargethispage{3\baselineskip}
\bibliographystyle{IEEEtran}
\bibliography{refs}

\end{document}